\documentclass[10pt,twocolumn,letterpaper]{article}

\usepackage{iccv}
\usepackage{times}
\usepackage{epsfig}
\usepackage{graphicx}
\usepackage{amsmath}
\usepackage{amssymb}
\usepackage{color}

\usepackage[pagebackref=true,breaklinks=true,letterpaper=true,colorlinks,bookmarks=false]{hyperref}
\usepackage{enumitem}
\usepackage{mathrsfs}
\usepackage{booktabs}

\iccvfinalcopy 

\def\iccvPaperID{1033} 

\ificcvfinal\fi
\begin{document}
\title{General Dynamic Scene Reconstruction from Multiple View Video}
\author{Armin Mustafa  \hspace{.09\linewidth}   Hansung Kim   \hspace{.09\linewidth}   Jean-Yves Guillemaut  \hspace{.09\linewidth}  Adrian Hilton\\
CVSSP, University of Surrey, Guildford, United Kingdom\\
{\tt\small a.mustafa@surrey.ac.uk}
}
\maketitle
\begin{abstract}
\vspace{-0.4cm}
This paper introduces a general approach to dynamic scene reconstruction from multiple moving cameras without prior knowledge or limiting constraints on the scene structure, appearance, or illumination.
Existing techniques for dynamic scene reconstruction from multiple wide-baseline camera views primarily focus on accurate reconstruction in controlled environments, where the cameras are fixed and calibrated and background is known.  These approaches are not robust for general dynamic scenes captured with sparse moving cameras.  
Previous approaches for outdoor dynamic scene reconstruction assume prior knowledge of the static background appearance and structure.
The primary contributions of this paper are twofold: an automatic method for initial coarse dynamic scene segmentation and reconstruction without prior knowledge of background appearance or structure; and a general robust approach for joint segmentation refinement and dense reconstruction of dynamic scenes from multiple wide-baseline static or moving cameras.
%
%
Evaluation is performed on a variety of indoor and outdoor scenes with cluttered backgrounds and multiple dynamic non-rigid objects such as people.
Comparison with state-of-the-art approaches demonstrates improved accuracy in both multiple view segmentation and dense reconstruction. The proposed approach also eliminates the requirement for prior knowledge of scene structure and appearance. 
\end{abstract}
\vspace{-0.6cm}
\section{Introduction}
\vspace{-0.2cm}
Reconstruction of general dynamic scenes is motivated by potential applications in film and broadcast production together with the ultimate goal of automatic understanding of real-world scenes from distributed camera networks. \\
\begin{figure}[t]
\begin{center}
\includegraphics[height = 2.6cm]{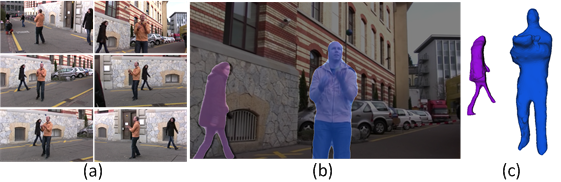}
\vspace{-0.1cm}
\caption{General dynamic scene reconstruction (a) Multi-view frames for Juggler dataset, (b) Segmentation of dynamic objects and (c) Reconstructed mesh}
\vspace{-0.6cm}
\label{fig:motivation}
\end{center}
\end{figure}
Over the past decades, effective approaches have been proposed to reconstruct dense dynamic shape from wide-baseline camera views in controlled environments with static backgrounds and illumination. A common assumption of widely used visual-hull based reconstruction approaches is prior foreground/background segmentation, which is commonly achieved using a uniform chroma-key color background or background image plate. Alternatively, multiple view stereo techniques have been developed which require a relatively dense camera network resulting in large numbers of cameras.

\vspace{-0.1cm}
Recent research has applied multiple view dynamic scene reconstruction techniques to less controlled outdoor scenes. Initial research focused on reconstruction in sports \cite{Guillemaut2010} exploiting known background images or the pitch color to obtain an initial segmentation. Extension to more general outdoor scenes \cite{UnstructuredVBR10,Kim2012,taneja2011modeling} uses prior reconstruction of the static geometry from images of the empty environment.
Research has also exploited strong prior models of dynamic scene structure such as people or used active depth sensors to reconstruct dynamic scenes. 

\vspace{-0.1cm}
This paper presents an approach for unsupervised dynamic scene reconstruction from multiple wide-baseline static or moving camera views without prior knowledge of the scene structure or background appearance. The input is a sparse set of synchronised  multiple view videos without segmentation. Camera extrinsics are automatically calibrated using scene features. An initial coarse reconstruction and segmentation of all dynamic scene objects is obtained from sparse features matched across multiple views. This eliminates the requirement for prior knowledge of the background scene appearance or structure. Joint segmentation and dense reconstruction refinement is then performed to estimate the non-rigid shape of dynamic objects at each frame. Robust methods are introduced to handle complex dynamic scene geometry in cluttered scenes from independently moving wide-baseline cameras views. The proposed approach overcomes constraints of existing approaches allowing the reconstruction of more general dynamic scenes. Results for a popular dataset, Juggler \cite{UnstructuredVBR10} captured with a network of moving handheld cameras are shown in Figure~\ref{fig:motivation}.
The contributions are as follows:
\begin{itemize}[topsep=0pt,partopsep=0pt,itemsep=0pt,parsep=0pt] 
\item Unsupervised dense reconstruction and segmentation of general dynamic scenes from multiple wide-baseline views.
\item Automatic initialization of dynamic object segmentation and reconstruction from sparse features. 
\item Robust spatio-temporal refinement of dense reconstruction and segmentation integrating error tolerant photo-consistency and edge information.
\end{itemize}
\vspace{-0.3cm}
%
\section{Related work}
\label{sec:survey}
\vspace{-0.2cm}
\subsection{Dynamic scene reconstruction}
\vspace{-0.2cm}
Research on multiple view dense dynamic reconstruction has primarily focused on indoor scenes with controlled illumination and backgrounds extending methods for multiple view reconstruction of static scenes \cite{Seitz06} to sequences \cite{Tung09}. 
In the last decade, focus has shifted to more challenging outdoor scenes captured with both static and moving cameras. 
Reconstruction of non-rigid dynamic objects in uncontrolled natural environments is challenging due to the scene complexity, illumination changes, shadows, occlusion and dynamic backgrounds with clutter such as trees or people.
Initial research focused on narrow baseline stereo  \cite{Cheng09, Larsen07} requiring a large number of closely spaced cameras for complete reconstruction of dynamic shape.
Practical reconstruction requires relatively sparse moving cameras to acquire coverage over large outdoor areas.  
A number of approaches for reconstruction of outdoor scenes require initial silhouette segmentation \cite{Wu2013,Kim2012,Li10, Guillemaut2010} to allow visual-hull reconstruction. Recent research has proposed reconstruction from a single handheld moving camera given a strong prior of bilayer segmentation \cite{Zhang11robustbilayer}. 
Bi-layer segmentation is used for depth-map reconstruction with the DAISY descriptor for matching \cite{Jiang12}, results are presented for handheld cameras with a relatively narrow baseline.

Pioneering research in general dynamic scene reconstruction from multiple handheld wide-baseline cameras \cite{UnstructuredVBR10, taneja2011modeling} exploited prior reconstruction of the background scene to allow dynamic foreground segmentation and reconstruction. This requires images of the environment captured in the absence of dynamic elements to recover the background geometry and appearance.

Most of these approaches to general dynamic scene reconstruction fail in case of complex (cluttered) scenes captured with moving cameras. 
These approaches either work for static/indoor scenes or exploit strong prior assumptions like silhouette information, known background or scene structure. Our aim is to perform dense reconstruction of dynamic scene automatically without any prior knowledge of background or segmentation of dynamic object.
\vspace{-0.12cm}
\subsection{Joint segmentation and reconstruction}
\vspace{-0.21cm}
Segmentation from multiple wide-baseline views has been proposed by exploiting appearance similarity \cite{Djelouah15, Lee11, Zeng04}. These approaches assume static backgrounds and different colour distributions for the foreground and background \cite{Sarim11, Djelouah15} which limits applicability for general scenes. In contrast to overcome these limitations, the proposed approaches initialised the foreground object segmentation from wide-baseline feature correspondence followed by joint segmentation and reconstruction.

Joint segmentation and reconstruction methods incorporate estimation of segmentation or matting with reconstruction to provide a combined solution. 
The first multi-view joint estimation system was proposed by Szeliski et al.\cite{Szeliski99} which used iterative gradient descent to perform an energy minimization.
A number of approaches were introduced for joint formulation in static scenes and one recent work used training data to classify the segments \cite{Zach13joint3d}. The focus shifted to joint segmentation and reconstruction for rigid objects in indoor and outdoor environment. Approaches used a variety of techniques like patch based refinement \cite{Shin2013, Ozden07} and fixation of cameras on the object of interest \cite{Campbell201014}.

Practical application of joint estimation requires these approaches to work on non-rigid objects like humans with clothing. Recent work proposed joint reconstruction and segmentation on monocular video achieving semantic segmentation of scene but does not work with dynamic objects \cite{Abhijit14}.
A multi-layer segmentation and reconstruction approach was proposed for sports data and indoor sequences \cite{Guillemaut2010} for multi-view videos. The algorithm used visual hull as a prior obtained from segmentation of the dynamic objects. The visual hull was optimized by combination of photo-consistency, silhouette, color and sparse feature information in an energy minimization framework to improve the segmentation and reconstruction quality. Although structurally similar to our approach it requires a background plate (assumed unknown in our case) as a prior to estimate the initial visual hull by background subtraction. The probabilistic color models of foreground and background are also used for optimization. A quantitative evaluation of state-of-the-art techniques for reconstruction from multiple views was presented by \cite{Seitz06}. These methods are able to produce high quality results, but rely on good initializations and strong prior assumptions.
\begin{figure*}[t]
\begin{center}
\includegraphics[height =4.5cm]{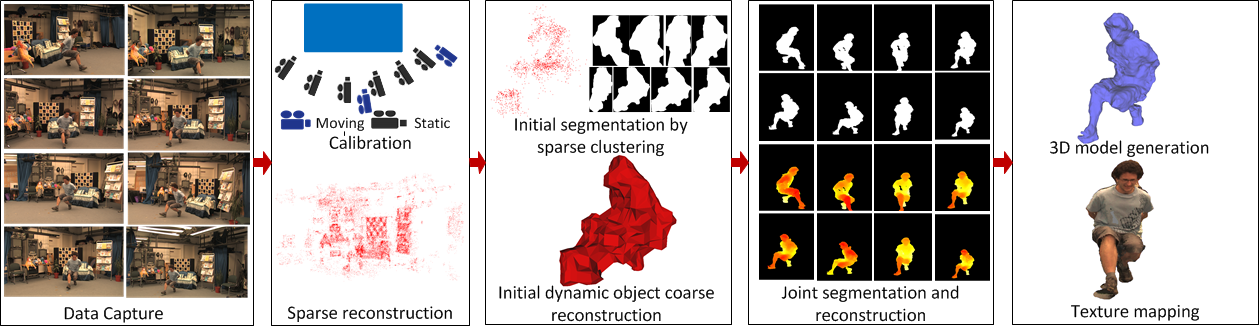}
\caption{Dense dynamic reconstruction framework}
\vspace{-1cm}
\label{fig:flow}
\end{center}
\end{figure*}

Image-based 3D dynamic scene reconstruction without a prior model is a key problem in computer vision. This research aims to overcome the limitations of the discussed approaches enabling robust wide-baseline multiple view reconstruction of general dynamic scenes without prior assumptions on scene appearance, structure or segmentation of the moving objects. The approach identifies and obtains an initial coarse reconstruction of dynamic objects automatically which is then refined using geometry and appearance cues in an optimization framework. The approach is a significant development over existing approaches as it works for the scenes captured only with moving cameras with unknown background and structure. Existing state-of-the-art techniques has not addressed this problem until now.
%
\vspace{-0.4cm}
\section{Overview}
\label{sec:method}
\vspace{-0.3cm}
%
The motivation of our work is to obtain automatic dense reconstruction and segmentation of complex dynamic scenes from multiple wide-baseline camera views without restrictive assumptions on scene structure or camera motion. 
The proposed approach estimates per-pixel dense depth with respect to each camera view of the observed moving non-rigid objects in the scene.  View-dependent depth maps are then fused to obtain a reconstruction for each dynamic object. An overview of the approach is presented in Figure \ref{fig:flow} and consists of the following stages:\\
\textbf{Data Capture:}
The scene is captured using multiple synchronised video cameras separated by wide-baseline. \\
\textbf{Calibration and sparse reconstruction:}
The intrinsics are assumed to be known for the static cameras and extrinsics are calibrated using Fundamental matrix estimation for pairs of images followed by bundle adjustment. Moving cameras are calibrated automatically using multi-camera calibration \cite{ImreGH11}.
A sparse 3D point-cloud is then reconstructed from wide-baseline feature matches.  \\
\textbf{Initial dynamic object segmentation and reconstruction:} 
Automatic initialisation is performed without prior knowledge of the scene structure or appearance to obtain an initial approximation for each dynamic object. 
Dynamic objects are segmented from the sparse 3D point cloud by combining optic flow with 3D clustering (section \ref{sec:Sparseclustering}). \\
\textbf{Joint segmentation and reconstruction for each dynamic object:}  The initial coarse reconstruction is refined for each dynamic object through joint optimisation of shape and segmentation using a robust cost function for wide-baseline matching.
View-dependent optimisation of depth is performed with respect to each camera which is robust to errors in camera calibration and initialisation. 
This gives a set of dense depth maps for each dynamic object.\\
\textbf{3D model generation and texture mapping:} A single 3D model for each dynamic object is obtained by fusion of the view-dependent depth maps using Poisson surface reconstruction \cite{Kazhdan2006}.
Surface orientation is estimated based on neighbouring pixels. 
Projective texture mapping is then performed for free-viewpoint video rendering.\\
\textbf{Dense reconstruction of sequence:} The process above is repeated for the entire sequence for all dynamic objects.

The proposed approach enables automatic reconstruction of all dynamic objects in the scene as a 4D mesh sequence. Subsequent sections present the novel contributions of this work in initialisation and refinement to obtain a dense reconstruction. The approach is demonstrated to outperform previous approaches to dynamic scene reconstruction and does not require prior knowledge of the scene structure. 
\vspace{-0.3cm}
\section{Initial dynamic object reconstruction}
\label{sec:Sparseclustering}
\vspace{-0.2cm}
For general dynamic scene reconstruction, we need to reconstruct and segment the dynamic objects in the scene at each frame instead of whole scene reconstruction for computational efficiency and to avoid redundancy. This requires an initial coarse approximation for initialisation of a subsequent refinement step to optimise the segmentation and reconstruction with respect to each camera view. We introduce an approach based on sparse point cloud clustering and optical flow labelling. This approach is robust to scene clutter in the 3D point cloud segmentation and partial segmentation of the dynamic object using optic flow due to partial motion or correspondence failure. Initialisation gives a complete coarse segmentation and reconstruction of each dynamic object for subsequent refinement. The optic flow and cluster information for each dynamic object helps us to retain same labels for the entire sequence. 
\vspace{-0.2cm}
\subsection{Sparse point cloud clustering}
\vspace{-0.2cm}
Feature detection is performed on all the multi-view images \cite{Mustafa15}. This is followed by SIFT descriptor based feature matching \cite{Lowe04} to obtain sparse reconstruction of the scene using the calibration information \cite{ImreGH11} for each time instant. This representation of the scene is processed to remove outliers using the point neighbourhood statistics to filter outlier data \cite{RusuDoctoralDissertation}. To retrieve the sparse features corresponding to the dynamic objects from the sparse reconstruction of the scene, we classify this representation into clusters followed by optical flow labelling.  
Data clustering approach is applied based on the 3D grid subdivision of the space using an octree data structure in Eucildean space. In a more general sense, nearest neighbors information is used to cluster, that is essentially similar to a flood fill algorithm \cite{RusuDoctoralDissertation}. 
We choose this because of its computational efficiency and robustness. The approach allows unsupervised segmentation of dynamic objects and is proved to work well for cluttered and general outdoor scenes as shown in Section \ref{sec:results}.
\begin{figure}[t]
\begin{center}
\includegraphics[height = 2.8cm]{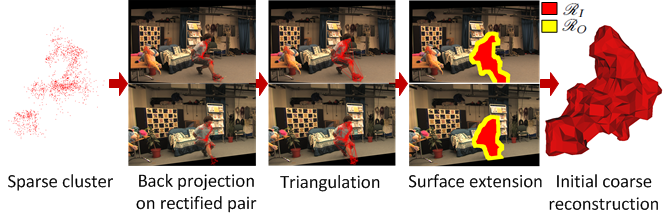}
\caption{Initial coarse reconstruction of the dynamic object in Odzemok dataset}
\vspace{-0.8cm}
\label{fig:ICR}
\end{center}
\end{figure}
\begin{figure}[t]
\begin{center}
\includegraphics[height = 2.8cm]{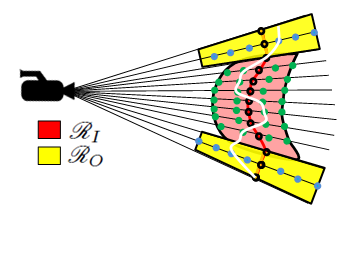}
\vspace{-0.6cm}
\caption{Initial coarse reconstruction: White line represents the actual surface, Depth labels are represented as circles; blue circles depict depth labels in $\mathscr{D}_{O}$, green circles depict depth labels in $\mathscr{D}_{I}$ and black circles depict the initial surface estimate.} 
\vspace{-0.55cm}
\label{fig:ICR2}
\end{center}
\end{figure}
\vspace{-0.4cm}
\subsection{Coarse scene reconstruction}
\vspace{-0.2cm}
Dynamic elements of the scene are identified by performing optical flow \cite{Zach07} on consecutive frames for a single view of each cluster. For each cluster the optimal camera view is dynamically selected to maximise visibility based on the sparse dynamic feature points at each frame. This allows efficient selection of the best view for optical flow. Optical flow is used to assign a unique label for each dynamic cluster throughout the sequence. If an object does not move between two consecutive time instants the reconstruction from this previous frame is retained. This limits the dynamic scene reconstruction to objects which have moved between frames reducing computational cost.\\
The process to obtain the coarse reconstruction is shown in Figure \ref{fig:ICR} and \ref{fig:ICR2}. The sparse representation of dynamic element is back-projected on the rectified image pair for each view. Delaunay triangulation \textcolor{red}{\cite{Fortune97}} is performed on the set of back projected points for each cluster on one image and is propagated to the second image using the sparse matched features. Triangles with edge length greater than the median length of edges of all triangles are removed. For each remaining triangle pair direct linear transform is used to estimate the affine homography \cite{Mustafa14}. Displacement at each pixel within the triangle pair is estimated by interpolation to get an initial dense disparity map for each cluster in the 2D image pair labelled as $\mathscr{R}_{I}$ depicted in red in Figure \ref{fig:ICR} and \ref{fig:ICR2}. 
The region $\mathscr{R}_{I}$ does not ensure complete coverage of the object, so we extrapolate this region to obtain a region $\mathscr{R}_{O}$ (shown in yellow) in 2D by $5\%$ of the average distance between the boundary points($\mathscr{R}_{I}$) and the centroid of the object. We assume that the object boundaries lie within the initial coarse estimate and depth at each pixel for the combined regions may not be accurate. Hence, to handle these errors in depth we add volume in front and behind of the projected surface by an error tolerance (calculated experimentally), along the optical ray of the camera. This tolerance may vary if a pixel belongs to $\mathscr{R}_{I}$ or $\mathscr{R}_{O}$ as the propagated pixels of the extrapolated regions ($\mathscr{R}_{O}$) may have a high level of errors compared to error at the points from sparse representation ($\mathscr{R}_{I}$) requiring a comparatively higher tolerance. The calculation of threshold depends on the capture volume of the datasets and is set to $1\%$ of the capture volume for $\mathscr{R}_{O}$ and half the value for $\mathscr{R}_{I}$. This volume in 3D corresponds to our initial coarse reconstruction of the dynamic object and enables us to remove the dependency of the existing approaches on background plate and visual hull estimates. This process of cluster identification and coarse reconstruction can be performed for multiple dynamic objects in the complex general environments. Initial dynamic object segmentation using point cloud clustering and coarse segmentation is insensitive to parameters. Throughout this work the same parameters are used for all datasets.
\vspace{-0.25cm}
\section{Joint segmentation and reconstruction}
\label{sec:optimize}
\vspace{-0.2cm}
\subsection{Problem statement}
\label{subsec:problem}
\vspace{-0.2cm}
In this section our aim is to refine the depth of the initial coarse reconstruction estimate of each dynamic object.
We aim to assign an accurate depth value to each pixel $p$ from a set of depth values $\mathscr{D} = \left \{ d_{1},...,d_{\left|\mathscr{D} \right|-1} , \mathscr{U} \right \}$. Each $d_{i}$ is obtained by sampling the optical ray from the camera and $\mathscr{U}$ is an unknown depth value to handle occlusions and to refine object segmentation. We assume that the depth of a particular pixel lies within the given threshold around the initial estimate as depicted in Figure \ref{fig:ICR2} and varies depending upon the regions $\mathscr{R}_{I}$ or $\mathscr{R}_{O}$. Hence we divide our depth labels in two sets, one for the region $\mathscr{R}_{I}$ ($\mathscr{D}_{I}$) and other for $\mathscr{R}_{O}$ ($\mathscr{D}_{O}$) such that $\left | \mathscr{D}_{I} \right | < \left | \mathscr{D}_{O} \right |$.
\vspace{-0.1cm}
\subsection{Proposed approach}
\label{sec:dense}
\vspace{-0.2cm}
We formulate the computation of depth at each point as energy minimization of the cost function defined in Eq. (\ref{eq:costfunction}). This equation is specifically designed to refine the reconstruction and segmentation and is used to estimate a view-dependent depth map for each dynamic object with respect to each camera. \\
$ E(d) = \lambda _{data}E_{data}(d) + \lambda _{contrast}E_{contrast}(d)  +$
\vspace{-0.2cm}
\begin{equation} \label{eq:costfunction}
\hspace{-2.8cm} \lambda _{smooth}E_{smooth}(d) 
\end{equation}
where, $d$ is the depth at each pixel for our dynamic object for the region $\mathscr{R}_{I}$ + $\mathscr{R}_{O}$ and can be assigned $\mathscr{U}$ to refine object segmentation. The equation consist of three terms: the data term is for the photo-consistency scores, the smoothness term is to avoid sudden peaks in depth and maintain the consistency and the contrast term is to identify the object boundaries. Data and smoothness terms are common to solve reconstruction problems \cite{Bleyer11} and the contrast term is used for segmentation \cite{Kolmogorov06}.
\vspace{-0.4cm}
\subsubsection{Matching term}
\vspace{-0.2cm}
To measure photo-consistency, we use a data term measure based on NCC recently proposed in \cite{Hu12}. They suggests this to be the best photo-consistency measure for wide baseline multi-view datasets because of its ability to obtain a high number of correct matches and preserve boundaries.\\
$ E_{data}(d) = \sum_{p\in \mathscr{P}} e_{data}(p, d_{p}) = $
\vspace{-0.2cm}
\begin{equation} \label{eq:matching1}
\hspace{-0.75cm}
\begin{cases}
    M(p, q)  = \sum_{i \in \mathscr{C}_{k}}m(p,q) ,& \text{if } d_{p}\neq \mathscr{U}\\
    M_{\mathscr{U}}, & \text{if } d_{p} = \mathscr{U}\\
\end{cases}
\vspace{-0.2cm}
\end{equation}
where $\mathscr{P}$ is the 4-connected neighbourhood of pixel $p$, $M_{\mathscr{U}}$ is the fixed cost of labelling a pixel unknown and $q = \Pi(p, d_{p})$ denotes the projection of the hypothesised point $P$ in an auxiliary camera where $P$ is the coordinates of $3D$ point along the optical ray passing through pixel $p$ located at a distance $d_{p}$ from the reference camera. $\mathscr{C}_{k}$ is the set of $k$ most photo-consistent pairs with reference camera.\\
For textured scenes NCC over a squared window is a common choice \cite{Seitz06}. The NCC values range from -1 to 1 which are then mapped to non-negative values by using the function $1 - NCC$. A maximum likelihood measure\cite{Larry92} is used in this function for confidence value calculation between the center pixel $p$ and the other pixels $q$ and is based on the survey on confidence measures for stereo \cite{Hu12}. The measure is defined as:
\vspace{-0.4cm}
\begin{equation} \label{eq:matching4}
m(p,q) = \frac{exp\tfrac{c_{min}}{2\sigma_{i}^{2}}}{\sum_{(p,q) \in \mathscr{N}} exp\tfrac{-(1-NCC(p,q))}{2\sigma_{i} ^{2}}}
\vspace{-0.2cm}
\end{equation}
where $\sigma_{i} ^{2}$ is the noise variance for each auxiliary camera $i$; this parameter was fixed to $0.3$. $\mathscr{N}$ denotes the set of interacting pixels in $\mathscr{P}$. $c_{min}$ is the minimum cost for a pixel obtained by evaluating the function $(1 - NCC(.,.))$ on a $15 \times 15$ window.
\vspace{-0.4cm}
\subsubsection{Contrast term}
\vspace{-0.2cm}
Segmentation boundaries in images tend to align with contours of high contrast and it is desirable to represent this as a constraint in stereo matching. A consistent interpretation of segmentation-prior and contrast-likelihood is used from \cite{Kolmogorov06}. We used a modified version of this interpretation in our formulation to preserve the edges by using Bilateral filtering \cite{Tomasi98} instead of Gaussian filtering.
\vspace{-0.3cm}
\begin{equation} \label{eq:contrast1}
E_{contrast} =  \sum_{p,q \in \mathscr{N}} e_{contrast}(p,q)
\end{equation}
\vspace{-0.3cm}
\begin{equation} \label{eq:contrast2}
e_{contrast}(p,q)=  
\begin{cases}
   0, & \hspace{-2.8cm} \text{if } (d_{p} =  d_{q} = \mathscr{U}) \text{ or } \\
   & \hspace{-2.5cm} (d_{p} =  d_{q} \neq \mathscr{U})\\
   \frac{1}{1+\epsilon }( \epsilon + exp^{-C(p,q)}),  & \text{otherwise}
\end{cases}
\end{equation}
$\left \| \cdot  \right \|$ is the $L_{2}$ norm and $\epsilon = 1$. The simplest choice for $C(p,q)$ would be the squared Euclidean color distance between intensities at pixel $p$ and $q$ as used in \cite{Guillemaut2010}. We propose a term for better segmentation as  $C(p,q) = \frac{\left \| B(p) - B(q) \right \|^{2}}{2 \sigma _{pq}^{2} d_{pq}^{2} }$ where $B(.)$ represents the bilateral filter, $d_{pq}$ is the Euclidean distance between $p$ and $q$, and $\sigma _{pq} = \left \langle\frac{\left \| B(p) - B(p)\right \|^{2}}{d_{pq}^{2}}\right\rangle$
This term enables to remove the regions with low photo-consistency scores and weak edges and thereby helps in estimating the object boundaries.
\vspace{-0.5cm}
\subsubsection{Smoothness term}
\vspace{-0.2cm}
This term is inspired by \cite{Guillemaut2010} and it ensures the depth labels vary smoothly within the object reducing noise and peaks in the reconstructed surface. This is useful when the photo-consistency score is low and insufficient to assign depth to a pixel.
\vspace{-0.5cm}
\begin{equation} \label{eq:smooth}
E_{smooth}(d) = \sum_{(p,q)\in \mathscr{N}} e_{smooth}(d_{p}, d_{q})\\
\end{equation}
\vspace{-0.5cm}
\begin{equation}
e_{smooth}(d_{p}, d_{q})=
\begin{cases}
    min(\left | d_{p} - d_{q} \right |, d_{max}),& \text{if } d_{p},d_{q}\neq \mathscr{U}\\
    0,              & \text{if } d_{p},d_{q} = \mathscr{U}\\
    d_{max},  & \text{otherwise}
\end{cases}
\end{equation}
$d_{max}$ is set to 50 times the size of the depth sampling step defined in Section \ref{subsec:problem} for all datasets.
\vspace{-0.3cm}
\subsection{Optimization of Reconstruction and Segmentation}
\vspace{-0.2cm}
The energy minimization for Eq. (\ref{eq:costfunction}) is performed by using the $\alpha$-expansion move algorithm from \cite{Boykov01}. We choose graph cuts because of its strong optimality properties over belief propagation \cite{Tappen03}. Graph-cut using the min-cut/max-flow algorithm is used to obtain a local optimum \cite{Kolmogorov04}. The  $\alpha$-expansion for a pixel $p$ is performed by iterating through the set of depth labels $\mathscr{D}_{I}$, if $p \in \mathscr{R}_{I}$ and $\mathscr{D}_{O}$, if $p \in \mathscr{R}_{O}$. Convergence is achieved after 4 or 5 iterations. A final model is obtained by merging the view-dependent depth representations through the Poisson surface reconstruction algorithm as explained in Section \ref{sec:method}.
\begin{figure*}[t]
\begin{center}
\fbox{\includegraphics[width = 15 cm]{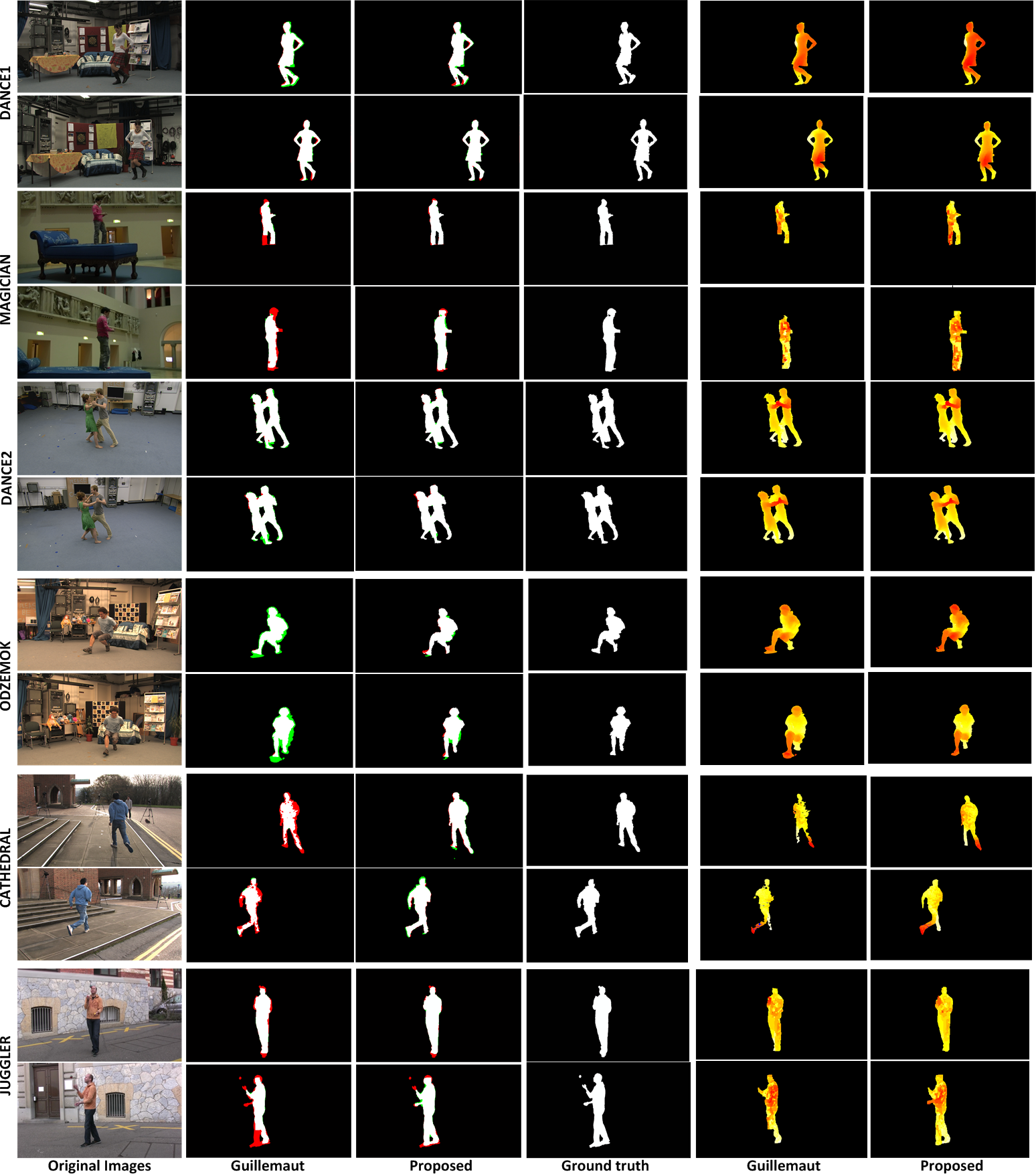}}
\caption{Results for a pair of images from each dataset:  $2^{nd} - 4^{th}$ column: Segmentation (Red represents true negatives and green represents false positives compared to the ground truth) and $5^{th} - 6^{th}$ column: Depth}
\vspace{-0.9cm}
\label{fig:result}
\end{center}
\end{figure*}
\vspace{-0.4cm}
\section{Results and Evaluation}
\label{sec:results}
\vspace{-0.2cm}
Evaluation is performed from publicly available research datasets: Indoor and Outdoor dataset with simple background (Dance2 and Cathedral), Indoor datasets with cluttered background (Odzemok and Dance1) (cvssp.org/cvssp3d) and Indoor and Outdoor datasets captured with moving handheld cameras (Magician and Juggler) \cite{UnstructuredVBR10}. The detailed characteristics of these datasets and the parameter settings for Eq.(\ref{eq:costfunction}) are summarised in Table \ref{dataset}. 
The framework explained in Section \ref{sec:method} is applied to all datasets, starting from sparse reconstruction followed by clustering and initial coarse reconstruction of dynamic objects which is then optimized using the proposed joint segmentation and reconstruction approach. Most existing methods do not perform simultaneous segmentation and reconstruction, therefore the method is compared to two state of the art approaches Furukawa and Ponce \cite{Furukawa10} for wide-baseline reconstruction and Guillemaut and Hilton \cite{Guillemaut2010} for joint reconstruction and segmentation. Both of these approaches are top performers on the Middlebury for multi-view reconstruction of wide-baseline views \cite{Seitz06}.
\vspace{-0.8cm}
\subsection{Segmentation results}
\label{sec:segresults}
\vspace{-0.3cm}
The segmentation results from the proposed approach are compared against the segmentation from Guillemaut and Hilton \cite{Guillemaut2010} and the ground-truth. Ground truth is obtained by manually labelling the foreground for all datasets except Juggler and Magician where ground-truth is available online.\\
\textbf{Guillemaut  \cite{Guillemaut2010}}: This approach requires an initial coarse foreground segmentation retrieved by differencing against a static background plate to obtain a visual hull required as a prior for reconstruction. In the proposed approach we do not assume a known background allowing the use of moving cameras. We modified the Guillemaut method by assigning the coefficient of the color term to be zero because we assume no prior knowledge of the background and we initialized this approach using our initial coarse reconstruction instead of the visual hull.
\begin{table*}[t]
\begin{center}
\begin{tabular}{|c|c|c|c|c|c|c|c|} 
\hline
Dataset & Number of Cameras & Number of frames & Image resolution & Baseline & $\lambda _{data}$ & $\lambda _{smooth}$ & $\lambda _{contrast}$ \\
\hline
Dance1 & 8 (1 moving) & 250 & $1920 \times 1080$ & 15 degrees & 0.5 & 0.005& 1.0 \\
Magician & 6 (all moving) & 6900 & $960 \times 544$ & 40-55 degrees & 0.6 & 0.01 & 3.0  \\
Dance2 & 8 (all static) & 125 & $1920 \times 1080$ & 45 degrees & 0.5 & 0.005 & 1.0 \\
Odzemok & 8 (2 moving) & 250 & $1920 \times 1080$ & 15 degrees & 0.5 & 0.005& 1.0 \\
Cathedral & 8 (all static) & 143 & $1920 \times 1080$ & 45 degrees & 0.6 & 0.01 & 5.0   \\
Juggler & 6 (all moving) & 3500 & $960 \times 544$ & 25-35 degrees & 0.6 & 0.01 & 5.0  \\
\hline
\end{tabular}
\end{center}
\vspace{-0.25cm}
\caption{Characteristics and parameter settings for datasets}
\vspace{-0.7cm}
\label{dataset}
\end{table*}
\vspace{-0.5cm}
\subsubsection{Qualitative results}
\vspace{-0.2cm}
The segmentation results for two frames from each dataset are shown in Figure \ref{fig:result}. Guillemaut requires accurate visual hull initialization, in this case the proposed coarse reconstruction is erroneous and far-away from the actual object boundaries as shown in Figure \ref{fig:ICR}. This results in less accurate segmentation compared to the proposed approach which disambiguates the problem by improving the contrast and data terms in the energy formulation. The data term removes the regions with very low photo-consistency and the contrast term introduces affinity towards strong edges of foreground. The artefacts with respect to ground truth in the proposed approach are from shadow areas and occlusions. 
\vspace{-0.4cm}
\subsubsection{Quantitative evaluation}
\vspace{-0.2cm}
To perform the quantitative evaluation of the segmentation we measured the $HitRatio$, $BkgRatio$ and $OverlapRatio$ as defined in \cite{Shin2013} against the ground truth pixels. The three criterion are defined as follows: \\
$HitRatio = \left | Result \bigcap GT \right | / \left | GT \right |\\
BkgRatio = \left | Result - GT \right | / \left | Result \right |$
\vspace{-0.4cm}
\begin{equation} \label{eq:overlapratio}
\hspace{-0.45cm}OverlapRatio =  \left | Result \bigcap GT \right | / \left | Result \bigcup GT \right |
\vspace{-0.2cm}
\end{equation}
The results are shown in Table \ref{segmentation} for all the dataset. The comparison parameters are averaged over the entire sequence to ensure the accuracy of the result. Higher hit, overlap ratio and lower background ratio represents better segmentation. The $HitRatio$ is the ratio of true positive in the result with the ground truth. The $OverlapRatio$ is the ratio of true positives in the result with the sum of result and ground truth. The ratios for the proposed approach are higher than Guillemaut for all the datasets, generally much higher for more complex datasets like outdoor scenes or scenes captured with only handheld moving cameras. This demonstrates the robustness of the proposed approach to general dynamic scene segmentation compared to Guillemaut as seen in Figure \ref{fig:result}. The $BkgRatio$ measures the proportion of result which actually belongs to background i.e. false positives in the segmentation. In case of Guillemaut this value is higher as compared to the proposed approach for most of the datasets. To conclude the segmentation obtained by the proposed approach vs. a state-of-the-art technique which assumes static cameras and a known background plate is better in quality with higher hit, overlap ratio and lower background ratio.
\vspace{-0.3cm}
\begin{table*}
\begin{center}
\begin{tabular}{lllllllllllll}
\toprule 
 Criteria & \multicolumn{2}{c}{Dance1} &  \multicolumn{2}{c}{Magician} & \multicolumn{2}{c}{Dance2}& \multicolumn{2}{c}{Odzemok} & \multicolumn{2}{c}{Cathedral} & \multicolumn{2}{c}{Juggler}\\

    & Ours & Guill.
    & Ours & Guill. 
    & Ours & Guill.
    & Ours & Guill.
    & Ours & Guill.
    & Ours & Guill.   \\
    \midrule
   HitRatio  & \textbf{0.995}   &0.993   & \textbf{0.887} & 0.663 &  \textbf{0.994}  & 0.992 & \textbf{0.899} & 0.895 & \textbf{0.891} & 0.796 & \textbf{0.879} & 0.646 \\
   BkgRatio  & \textbf{0.023}   &0.042  & 0.022 & \textbf{0.018} &  \textbf{0.020}  & 0.031 & \textbf{0.381} & 0.507 & 0.021 & \textbf{0.015} & \textbf{0.025} & 0.038\\
   Overlap  & \textbf{0.947}   &0.928 & \textbf{0.855} & 0.595  &  \textbf{0.963}  & 0.941 & \textbf{0.611} & 0.469 & \textbf{0.849} & 0.745 & \textbf{0.841} & 0.577 \\
    \bottomrule
\end{tabular}
\end{center}
\vspace{-0.2cm}
\caption{Segmentation performance comparison for all datasets (best for each dataset is highlighted in bold) (Guill. depicts Guillemaut)}
\vspace{-0.6cm}
\label{segmentation}
\end{table*}
\begin{figure}[b]
\begin{center}
\includegraphics[width = 8.3cm]{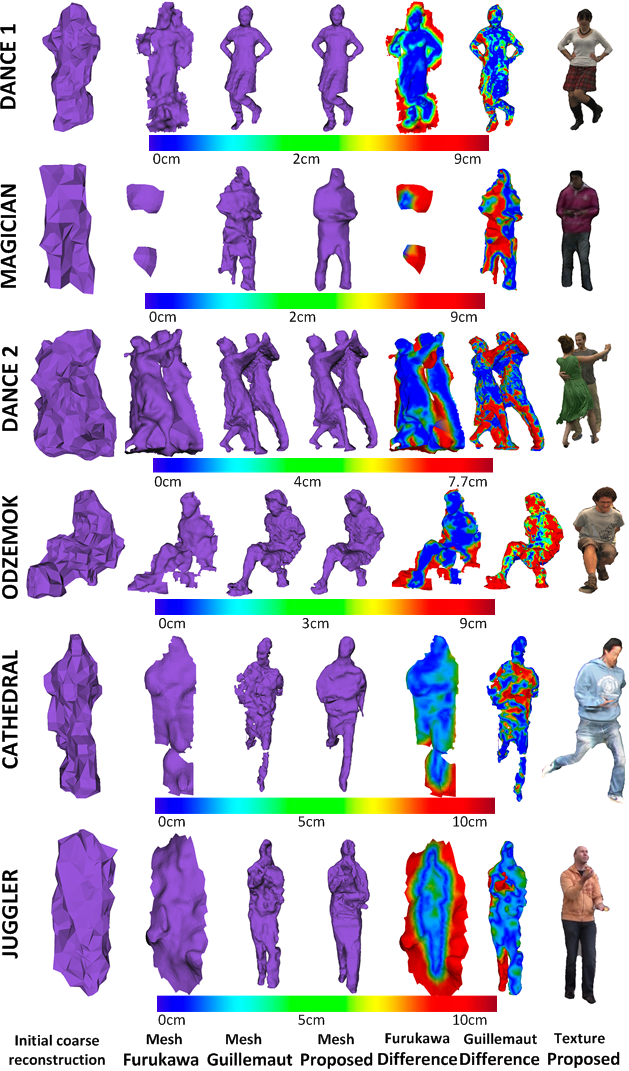}
\caption{Results for each dataset: $1^{st} - 4^{th}$ column: Meshes and $5^{th} - 6^{th}$ column: Difference meshes against proposed approach with color coded error in cms and $7^{th}$ is textured mesh.}
\vspace{-0.5cm}
\label{fig:meshes}
\end{center}
\end{figure}
\subsection{Reconstruction results}
\vspace{-0.2cm}
We have compared our results with Guillemaut (Section \ref{sec:segresults}) and 
\textbf{Furukawa \cite{Furukawa10}}: This represents a state-of-the-art multi-view wide-baseline stereo approach. Furukawa \cite{Furukawa10} does not refine the segmentation but gives a 3D point cloud which is converted into a mesh using Poisson surface reconstruction.
For fair comparisons all of the approaches are initialised with the same calibration and coarse reconstruction obtained using the method explained in Section \ref{sec:Sparseclustering}. 
\begin{figure}[b]
\begin{center}
\includegraphics[height = 3.5cm]{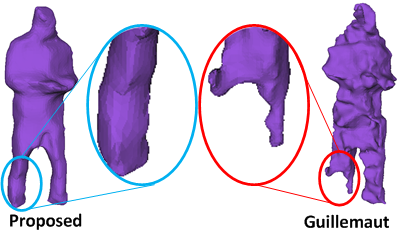}
\caption{Result for magician dataset}
\vspace{-0.5cm}
\label{fig:magician}
\end{center}
\end{figure}
\vspace{-0.3cm}
\subsubsection{Qualitative results}
\vspace{-0.2cm}
The depth maps for the proposed approach and Guillemaut are shown in Figure \ref{fig:result}. The consistency of depth maps in the case of the proposed approach are better because of the use of an improved data term for robustly matching between views and preserving edges.\\
The 3D models of the dynamic foreground obtained from the proposed approach are compared with Guillemaut and Furukawa in Figure \ref{fig:meshes} for all the datasets. For Magician dataset Furukawa gives very few points on a small part of the object in the reconstruction due to the complexity of the dataset. Results are compared closely with Guillemaut in Figure \ref{fig:magician}.
In Figure \ref{fig:meshes} the meshes obtained by Furukawa do not have clear boundaries because it is not designed to refine the segmentation of the object. The meshes obtained from the proposed approach are visibly more accurate compared to the other techniques especially in the case of outdoor datasets. Some errors in the mesh reconstruction are present due to camera noise, uniform textures and similarity to the background. 
Results for Juggler sequence are shown in Figure \ref{fig:sequence} and more results are available in supplementary material and video.
\begin{figure}[b]
\begin{center}
\includegraphics[width =8.5cm]{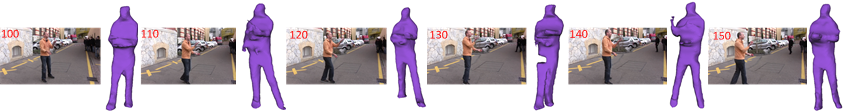}
\caption{Result for Juggler sequence: Original images from one view with frame numbers and mesh reconstructions alternatively}
\vspace{-0.5cm}
\label{fig:sequence}
\end{center}
\end{figure}
\vspace{-0.5cm}
\subsubsection{Quantitative evaluation}
\vspace{-0.2cm}
Due to the absence of ground-truth 3D models for the datasets the accuracy evaluation is limited to the qualitative analysis. In this section we compare the computational efficiency of different approaches against the proposed method. The run-time per frame is shown in Table \ref{time}. The speed of the proposed approach is slightly lower than Furukawa (which does not perform segmentation) and the improvement in the speed for the proposed approach is approximately 25\% as compared to Guillemaut.
\begin{table}[h]
\begin{center}
\begin{tabular}{|c|c|c|c|} 
\hline
Dataset & Furukawa\cite{Furukawa10} & Guillemaut\cite{Guillemaut2010} & Proposed \\
\hline
Dance1 & 326 s  & 448 s &  295 s \\
Magician & 311 s & 452 s & 377 s \\
Dance2 & 502 s & 655 s & 471 s  \\
Odzemok & 381 s & 498 s & 364 s \\
Cathedral & 525 s & 679 s & 501 s \\
Juggler & 399 s & 466 s & 374 s \\
\hline
\end{tabular}
\end{center}
\vspace{-0.2cm}
\caption{Comparison of computational efficiency for all datasets (time in seconds (s))}
\label{time}
\end{table}
\subsection{Limitations and Future work}
\vspace{-0.2cm}
The proposed approach reconstructs and segments multiple close objects as a single dynamic object. This is not a failure case, but it increases the overall computational time of general scene reconstruction. Secondly, the proposed technique does not handle textureless scenes due to the sparcity of 3D points and crowded scenes due to the failure of the clustering algorithm used for initialisation. We aim to handle these scenes in future, by inclusion of full scene reconstruction from the sequence.
\vspace{-0.2cm}
\section{Conclusion}
\label{sec:conclusion}
\vspace{-0.2cm}
This paper introduced a novel technique to automatically segment and reconstruct dynamic objects captured from multiple moving cameras in general dynamic uncontrolled environments without any prior on background appearance or structure. The proposed automatic initialization was used to identify and initialize the segment and reconstruction of multiple dynamic objects. The initial coarse approximation is refined using a a joint view-dependent optimisation of segmentation and reconstruction by a view-dependent graph-cut optimization using the photo-consistency and contrast cues from wide-baseline images.

Unlike previous method the proposed approach allows unsupervised reconstruction 
without prior information on scene appearance or structure. The segmentation and reconstruction accuracy are significantly improved over previous methods allows application to more general dynamic scenes. Tests on challenging datasets demonstrate improvements in quality of reconstruction and segmentation compared to state-of-the-art methods. \\

\noindent \textbf{Acknowledgements}\\
This research was supported by the European Commission, FP7 Intelligent Management Platform for Advanced Real-time Media Processes project (grant 316564).
{\small
\bibliographystyle{ieee}
\bibliography{egbib}
}
\end{document}